\documentclass[twoside,11pt]{article}
\usepackage{jmlr2e}
\usepackage{amsmath}
\usepackage{graphics}
\usepackage{graphicx}
\usepackage{txfonts}
\usepackage{subfigure}
\usepackage{algorithm}
\usepackage{algorithmic}
\usepackage{caption}
\usepackage{float}
\usepackage{threeparttable}
\usepackage{url}
\usepackage{multirow}
\usepackage{diagbox}
\usepackage{amsmath}
\usepackage{graphics}
\usepackage{graphicx}
\usepackage{txfonts}
\usepackage{subfigure}
\usepackage{algorithm}
\usepackage{algorithmic}
\usepackage{caption}
\usepackage{float}
\usepackage{threeparttable}
\usepackage{url}
\usepackage{multirow}
\usepackage{diagbox}
\usepackage{amssymb}
\usepackage[utf8]{inputenc} 
\usepackage[T1]{fontenc}    
\usepackage{hyperref}       
\usepackage{url}            
\usepackage{booktabs}       
\usepackage{amsfonts}       
\usepackage{nicefrac}       
\usepackage{microtype}      
\usepackage{xcolor}         
\newtheorem{Theorem}{Theorem}
\newtheorem{Lemma}{lemma}
\newtheorem{Definition}{Definition}

\jmlrheading{1}{2000}{1-48}{4/00}{10/00}{Marina Meil\u{a} and Michael I. Jordan}
\ShortHeadings{Gaussian mixture modeling of nodes in Bayesian network according to maximal parental cliques}{Gaussian mixture modeling of nodes in Bayesian network according to maximal parental cliques}
\firstpageno{1}

\begin{document}
\title{Gaussian mixture modeling of nodes in Bayesian network according to maximal parental cliques}

\author{\name Yiran Dong \email 22035082@zju.edu.cn \\
       \addr School of Mathematical Sciences\\
       Zhejiang University\\
       Hangzhou 310027, China.
       \AND
       \name Chuanhou Gao \email gaochou@zju.edu.cn \\
       \addr School of Mathematical Sciences\\
       Zhejiang University\\
       Hangzhou 310027, China.}

\editor{}

\maketitle

\begin{abstract}
To capture the close relevance among nodes in Bayesian network, we take every clique in the network as a cluster, and construct Gaussian mixture model (GMM) on each node according to branches of its maximal parental cliques (MPCs). The definition and finding algorithm for MPCs are presented, accordingly. For the model of GMM-MPC, we targetedly propose an optimization algorithm to train the model parameters. Along with these algorithms, some theoretical analysis are further made to establish strong support. At the end, in experiments we use three public data sets to verify the effectiveness of our proposed method.
\end{abstract}

\section{Introduction}
Bayesian network (BN) is a kind of probabilistic graphical models (PGMs) (\cite{koller2009probabilistic}) that uses a graph to represent the joint distribution of a data set and the conditional independence in this distribution. Mathematically, it can be viewed as a directed acyclic graph (DAG), denoted by $\mathcal{G}=(\mathcal V,\mathcal E)$ with $\mathcal V$ to indicate the set of vertexes or nodes and $\mathcal E$ to be the set of directed edges. Every node in the network is usually to represent a random variable while every directed edge measures conditional dependencies between the connected two nodes. Due to having clear structure and strong interpretability, BN has been widely used in many respects, including medical diagnosis (\cite{glymour2016causal}), product recommendation (\cite{ono2007context}), images and sentences generation (\cite{dethlefs2011combining,kingma2019introduction}), etc. 

As a graphic model, the performance of BN depends on both the graphic structure and the distribution used to model the nodes. A ``good" BN thus needs to learn the network structure and the distribution (including the distribution form and parameters), respectively. For the former, the learning algorithms include constraint-based algorithms (\cite{colombo2014order, spirtes2000causation}) that measure the conditional independencies through independence tests, score-based algorithms that use a score function to rank the graph and find the graph 
with highest score, such as greedy equivalence search (GES) (\cite{chickering2002optimal}) and greedy interventional equivalence search (GIES) (\cite{hauser2012characterization}), and their hybrid methods, like max-min hill climbing (MMHC) algorithm (\cite{tsamardinos2006max}). For the distributions used in BN, as far as continuous data is concerned, the most frequently-used one is the linear Gaussian distribution. Based on it, some improved versions and other distributions have been also reported for the purpose of strengthening the performance of BN. The well-known Variational AutoEncoder (\cite{kingma2013auto}) uses deep neural networks to model the mean of the linear Gaussian model. Harris and Drton (\cite{harris2013pc}) proposed the nonparanormal distribution instead, which uses different strictly increasing functions to act on the normal distribution as the new distribution. Song et al. (\cite{song2011kernel}) used nonparametric representation as distribution, and Radu-Stefan et al. (\cite{niculescu2006bayesian}) imposed expert knowledge on constraining the parameters of the linear Gaussian distribution. 

The above improvements on model strengthen greatly the generative and discriminative abilities of BN, but the resulting models have too complex structures that are difficult to understand so as to cause loss of interpretability. For this reason, we try to use the Gaussian mixture model (GMM, a weighted average of finite Gaussian distribution) (\cite{mclachlan1988mixture}) in BN, which has transparent meaning on every parameter and thus can keep interpretability. Naturally, it is not the first time to apply GMM to model BN. Roos et al. applied GMM on dynamic BN to predict the passenger flow (\cite{roos2017dynamic}), and Monti et al. used GMM on naive BN to perform classification task of discrete data (\cite{monti2013bayesian}). However, these applications rely on special graphic structure of BN, such as dynamic BN or naive BN, and moreover, for every node GMM takes all of its parent nodes as a branch and is equipped with the same number of branches. These fixed settings may restrict the performance of the PGM combing BN and GMM. We thus study a very general case with no limitations on the structure of BN or on the number of branches in GMM. To capture the close relations among nodes, we utilize the clique structure (a cluster of nodes with any two nodes connected) and model every node with GMM according to its maximal parental clique (MPC), referred to as GMM-MPC in the context. A new optimization algorithm called double iteration optimization (DIO) algorithm is further proposed to optimize GMM-MPC with conditional variables in every branch under equality constraint. We finally use three data sets to test our models and algorithms.  

The rest of the paper is organized as follows. Section 2 introduces some background knowledge about BN and GMM. Section 3 shows the modeling framework of GMM-MPC in BN, including definition of MPC, DIO algorithm and some theoretical analysis. Then, some experiments and discussions are given in Section 4. Finally conclusion and other thinking are presented in Section 5.





\section{Preliminaries}
In this section, we will make a brief introduction on BN and GMM.
\subsection{Bayesian network}
Consider a BN $\mathcal{G}=(\mathcal V,\mathcal E)$. If $X,Y\in\mathcal V$ and $X\to Y$, then we call $X$ is a parent of $Y$, denoted by $\mathbf {Pa}_Y=\{X\}$, and $Y$ is a child of $X$, denoted by $\mathbf {Ch}_X=\{Y\}$. There are three basic structure in BN: chain ($X\to Y\to Z$), fork ($X\gets Y\to Z$) and v-structure ($X\to Y\gets Z$), where $Z\in\mathcal V$. In the v-structure, the middle node is defined as collider. If the parents of collider are disconnected, we say this v-structure is immorality. However, if a v-structure is not immorality, it is a complete subgraph and all the nodes in it can be colliders, so we call collider we mean the collider in immoralities. 

For a set of nodes $C$, if any pair of nodes in $C$ have an edge, $C$ is called a clique. Further, if there is no clique $C_0$ in graph $\mathcal G$ such that $C\subset C_0$, then $C$ is the maximal clique in $\mathcal G$. If a graph is a clique, we call this graph complete graph. All the complete graphs who have same nodes are I-equivalent since there is no conditional independencies.

Different graphs can represent the same distribution as long as they have the same nodes and  conditional independencies, we call these DAG are I-equivalent\cite{chickering2013transformational}.

Thus most of the structure learning algorithms return a partially directed acyclic graph which can become all I-equivalent DAGs that represent the same distribution of data by changing all the undirected edges to arbitrary directed edges\cite{chickering2013transformational}.

The values of nodes only depend on the parents of them and some Gaussian noises. The decomposition of a joint distribution is
\begin{align*}P(X_1,X_2,...,X_n)=\prod_i^n P(X_i\ |\ \mathbf{Pa}_i).
\end{align*}
The most widely used distribution in BN is linear Gaussian , $P(X_i\ |\ \mathbf{Pa}_i)=\mathcal N(X_i\ |\ (\mathbf w)^T\mathbf p+b,\sigma^2)$ where $\mathbf p$ is the value of $\mathbf{Pa}_i$, and $\mathbf w$, $\mathbf b$ and $\sigma$ are learnable parameters. Different structures of graph have different decomposition, thus have different conditional independencies, so a good structure learning algorithm can definitely improve the generative ability of BN. We do our experiments based on different structure learning algorithms.

\subsection{Gaussian mixture model and expectation maximization algorithm}
GMM is the weighted average of finite Gaussian distribution, 
\begin{align} P(\mathbf x)=\sum_{k=1}^K \pi_i& \mathcal N\left(\mathbf x\ |\ \boldsymbol\mu_k,\boldsymbol\Sigma_k\right)
\end{align}

with restriction $\underset{k}{\sum}\pi_k=1$, where each Gaussian distribution is called a branch or component, and $\mu_k$, $\Sigma_k$ and $\pi_k$ are the mean, variance and coefficient in k-th cluster. If the data set appears to have more than one clusters and a single Gaussian model is hard to fit the joint distribution, GMM uses its every branch to approximate every cluster of data. (figure 1 (a))

We set the Lagrange function $-\underset{j=1}{\overset{N}{\sum}}\ln\left(P(\mathbf x_j)\right)-\lambda\left(\underset{k=1}{\overset{K}{\sum}}\pi_k-1\right)$ as the loss function where $\lambda$ is the Lagrange multiplier. Setting the derivatives of the loss function with respect to the $\boldsymbol\mu_k$, $\boldsymbol\Sigma_k$ and $\pi_k$ for k=1,2,...,K, we obtain $\lambda=N$ and
\begin{align} \boldsymbol\mu_k=\frac1{N_k}\sum^N_{j=1}\gamma_{jk} \mathbf x_j,\ \ \ \ \ \ \ 
\boldsymbol\Sigma_k=\frac1{N_k}\sum_{j=1}^N \gamma_{jk} \left(\mathbf x_j-\boldsymbol\mu_k\right)\left(\mathbf x_j-\boldsymbol\mu_k\right)^T,\ \ \ \ \ \ \ 
\pi_k=\frac{N_k}N,
\end{align}
where $\gamma_{jk}=\displaystyle\frac{\pi_k\mathcal N(\mathbf x_j\ |\ \boldsymbol\mu_k,\boldsymbol\Sigma_k)}{\underset{k=1}{\overset{K}{\sum}}\pi_k\mathcal N(\mathbf x_j\ |\ \boldsymbol\mu_k,\boldsymbol\Sigma_k)}$ and $N_k=\overset{N}{\underset{j=1}{\sum}}\gamma_{jk}$\cite{bishop2006pattern}. EM algorithm has E-step and M-step in every epoch, In E-step, EM takes the $\boldsymbol\mu^{(t-1)}_k$, $\boldsymbol\Sigma_k^{(t-1)}$ and $\pi_k^{(t-1)}$ from the last epoch to compute $\gamma_{jk}^{(t)}$ for $j=1,...,N$ and $k=1,...,K$, in M-step it uses the data and $\gamma_{jk}^{(t)}$ to compute $\boldsymbol\mu^{(t)}_i$ and $\pi_i^{(t)}$ in equation (2), and takes $\boldsymbol\mu^{(t)}_i$ and $\gamma_{jk}^{(t)}$ as inputs to obtain $\boldsymbol\Sigma_k^{(t)}$. EM algorithm continues this routine until loss function converges.

\section{GMM framework}
In this section, we will give the definition of MPC, the searching algorithm for it, and the Gaussian mixture modeling framework on node according to its MPCs. 
\subsection{MPC}
Intuitively, the connection among nodes in a clique of BN is stronger than that in other set of nodes, since there is an edge between any pair of nodes. Cliques tree inference (\cite{kjaerulff1998inference}) utilizes this point and takes each clique as a cluster, passing information from one clique to another to obtain the marginal distributions. Inspired by this operation, we also model BN by taking every clique as a cluster, and for every node in clique we construct a GMM with every branch corresponding to one of its MPCs. Here, we only consider the cliques in $\mathbf{Pa}_T$, the main reason of which is that the value of $T$ only depends on its parents. The MPC is defined as follows. 

\begin{Definition} Given a BN $\mathcal{G}=(\mathcal V,\mathcal E)$, for any $T\in \mathcal V$ if the clique $C\subseteq\mathbf{Pa}_T$ and no other $C_0\subseteq\mathbf{Pa}_T$ such that $C\subset C_0$, then $C$ is called a maximal parental clique of $T$.
\end{Definition}

To construct GMM for node $T$, it is necessary to find all of its maximal parental cliques. From the definition, we can find that if a clique $C\subset\mathbf{Pa}_T$ and $C\cup \{T\}$ is a maximal clique, then $C$ is a MPC of $T$. Namely, the MPCs of $T$ are in $\mathbf{Pa}_T$, which means that $\mathbf{Pa}_T$ can be written as $\mathbf{Pa}_T=\{\mathbf{Pa}^T_1,...,\mathbf{Pa}^T_{K}\}$ with $\mathbf{Pa}^T_k~(k=1,...,K)$ to represent every MPC.(figure 1 (b)) Clearly, $\mathbf{Pa}_T^k$ comes from a clique of $T$, while the elements in clique might belong to $\mathbf{Pa}_T$ and/or $\mathbf{Ch}_T$, so we introduce an important notation $\mathbf{PC}_T=\mathbf{Pa}_T \cup\mathbf{Ch}_T$ serving for developing algorithm to find the MPCs of node $T$. 

\begin{algorithm}[H]
\renewcommand{\algorithmicrequire}{\textbf{Input:}}
\renewcommand{\algorithmicensure}{\textbf{Output:}}
\caption{Find maximal parental cliques}
\label{alg1}
\begin{algorithmic}[1]
\REQUIRE $\mathcal V$, all $\mathbf{PC}$ of $\mathcal V$
\STATE $\mathbf{C}_T=\emptyset$;
\FOR{$\mathbf{PC}_T$ in all different arrangements}
\STATE $clique=\emptyset$;
\FOR{$X\in \mathbf{PC}_T$}
\IF{$clique$ is a subset of $\mathbf{PC}_X$ and $X$ is the parent of $T$}
\STATE $clique=clique\cup\{X\}$;
\ENDIF
\ENDFOR
\IF{$clique$ not in $\mathbf{C}_T$}
\STATE $\mathbf{C}_T=\mathbf{C}_T\cup\{clique\}$;
\ENDIF
\ENDFOR
\ENSURE $\mathbf{C}_T=\{\mathbf{Pa}^T_1,...,\mathbf{Pa}^T_{K}\}$.
\end{algorithmic}
\end{algorithm}

In \textbf{Algorithm 1}, the $\mathbf{PC}$ of all nodes in $\mathcal V$ can be obtained by graphic structure learning algorithms, like MMHC, etc. Line 2 goes through all arrangements of nodes in $\mathbf{PC}_T$ with the purpose of finding all combinations of $T$ and nodes in $\mathbf{PC}_T$ to avoid getting only one clique given a kind of arrangement. Line 5 depends on the fact that a clique after adding another new node $X$ is still a clique if and only if this clique is in $\mathbf{PC}_X$. Meanwhile, to ensure this clique is a MPC, we need to check every node in this clique is parent of $T$. However, it is still possible to get the same clique from different arrangements, so we check the repetition in Line 9. The output $\mathbf C_T=\{\mathbf{Pa}^T_1,...,\mathbf{Pa}^T_{K}\}$ is the set of all MPCs.





\subsection{GMM of node according to branches of its MPC}
\begin{figure*}
\centering
\subfigure[]
{
 	\begin{minipage}[b]{0.48\linewidth}
        \centering
        \includegraphics[scale=0.55]{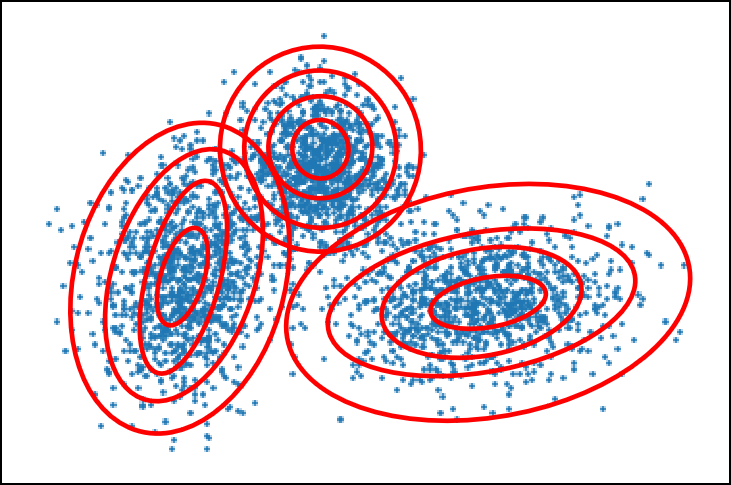}
    \end{minipage}
}
\subfigure[]
{
 	\begin{minipage}[b]{.45\linewidth}
        \centering
        \includegraphics[scale=0.5]{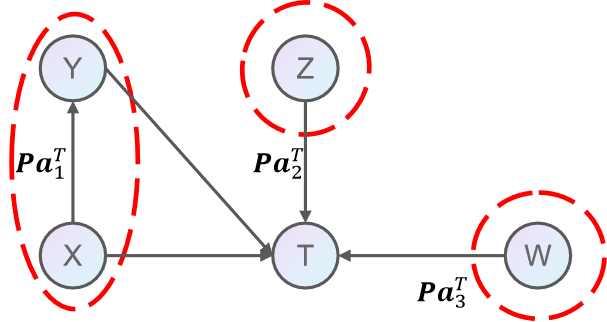}
    \end{minipage}
}

\caption{(a) GMM on 2-dimensional data set: GMM uses Gaussian distribution to fit each cluster; (b) GMM-MPC on BN: Nodes $X$, $Y$, $Z$, $W$ are parents of $T$, and $X$, $Y$, $T$ form a maximal clique, thus $X$, $Y$ form a MPC. In this example $T$ has three MPCs, we use Gaussian distribution as conditional distribution of $T$ given each MPC.}
\end{figure*}

Consider a data set $\mathbf D=\{\mathbf x_j\}_{j=1}^N$, where $N$ is the number of instances and $\mathbf x_j\in\mathbb{R}^n$. Let $X_i~(i=1,..,n)$ represent the $i$th feature variable of $\mathbf D$, corresponding to the $i$th node in the assigned BN $\mathcal{G}=(\mathcal V,\mathcal E)$. Then, the GMM for node $X_i$ is written as  
\begin{align}\label{eq:GMM}
P(X_i\ |\ \mathbf{Pa}_i)=\underset{k=1}{\overset{K_i}{\sum}} \pi_k^i\overline P^i_k(X_i\ |\ \mathbf{Pa}^i_k),
\end{align} 
where $K_i$ is the number of branches (every MPC acts for a branch), $\pi^i_k$ is the weighted coefficient of the $k$th MPC constrained by $\underset{k=1}{\overset{K_i}{\sum}} \pi_k^i=1$, $\mathbf{Pa}^i_k$ is the $k$th MPC of $X_i$, and $\overline P^i_k$ is the component distribution of $X_i$ given $\mathbf{Pa}^i_k$. By taking the conditional distribution in (\ref{eq:GMM}) as the posterior distribution, and the marginal distribution of node $X_i$ as the prior distribution, we have

\begin{Theorem}\label{thm1}
If the component distributions in (\ref{eq:GMM}) are conjugate distributions, then the mixture distribution there is also a conjugate distribution. Moreover, the joint distribution of the graph $\mathcal G$ is a mixture distribution i.e. $ P(\mathcal V)=\overset{K}{\underset{k=1}{\sum}}\pi_k \overline P_k(\mathcal V)$, where $K$ is the number of clusters of joint distribution and $\overset{K}{\underset{k=1}{\sum}}\pi_k=1$.
\end{Theorem}
Note that the left side of (\ref{eq:GMM}) is certainly seen as a mixture distribution, but the component distribution $\overline P$ in the right side can be also seen as a mixture model which has only one cluster. Similarly, it applies to the distributions of every subgraph. We put the detailed proof in the \textbf{Appendix}.

We say a graph $\mathcal G$ is a perfect map to data set $\mathbf D$ if $\mathcal G$ have all the conditional independencies in $\mathbf D$. Using \textbf{Theorem 1}, we have the following theorem.
\begin{Theorem}
Let $\mathcal G$ be a perfect map, then the marginal distribution of a node in $\mathcal G$ is a mixture distribution with more than one components if and only if the node is a collider or the descendant of a collider.
\end{Theorem}
In the \textbf{Appendix}, we prove only the collider and its descendants can have more than one MPCs. However, in the perfect graph the reverse is also true since if the coefficient of a MPC of the collider $T$ is zero, then the nodes in MPCs have no impact on $T$. That is to say they are not parents of $T$ which contradicts to the definition of the perfect map.

We consider the component distributions $\overline P$ as linear Gaussian model, similarly, we can obtain the Lagrange function about joint distribution of all nodes in $\mathcal V=\{X_1,...,X_n\}$
\begin{align} -\sum^N_{j=1}\sum^n_{i=1}\ln\left(\sum^{K_i}_{k=1}\pi_k^i\mathcal N\left(X_{ij}\ \big|\ \left(\mathbf w_k^i\right)^T\mathbf p^i_{jk}+b_k^i,\sigma^i_{k}\right)\right)-\sum^n_{i=1}\lambda_i\left(\sum^{K_i}_{k=1}\pi_k^i-1\right)
\end{align}
as loss function, where $X_{ij}$ is the value of i-th nodes in j-th data, $\mathbf p^i_{jk}$ is the $k$th MPC of $i$th node in $j$th data, $\mathbf w^i_k$, $b^i_k$, $\sigma^i_k$, $\pi^i_k$ are learnable parameters. Meanwhile, it has n equality constraints. The equation (4) can be seen as a likelihood function of n GMM-MPCs, but the mean of equation (1) are learnable parameters, and the mean of (4) contains some features of data.
\begin{Lemma} To achieve the minimum of equation (4), $\lambda_1=\lambda_2=...=\lambda_n=N$, and update equation in EM algorithm with respect to equation (4) is 
\begin{align}\left(\sum^{N}_{j=1}\gamma^i_{jk}\left(\mathbf p^i_{jk}\right)^T\mathbf p^i_{jk}\right)\mathbf w^i_k=&\sum^{N}_{j=1}\gamma^i_{jk}\left(X_{ij}-b^i_k\right)\left(\mathbf p^i_{jk}\right),\\
b^i_k=\frac{1}{N^i_k}\sum^{N}_{j=1}\gamma^i_{jk}\left(X_{ij}-\left(\mathbf w^i_{jk}\right)^Tp^i_{jk}\right),\ \ \ &\ \ \  \sigma^i_k=\frac{1}{N^i_k} \sum^{N}_{j=1}\gamma^i_{jk}\left(X_{ij}-\left(\mathbf w^i_k\right)^T\mathbf p^i_{jk}-b^i_k\right)^2,\\ \pi^i_k=\frac{N^i_k}{N},\ \ \ &\ \ \ N^i_k=\sum^{N}_{j=1}\gamma^i_{jk},
\end{align}
where $\gamma^i_{jk}=\displaystyle\frac{\pi^i_k\mathcal N\left(X_{ij}\ \big|\ \left(\mathbf w_k^i\right)^T\mathbf p^i_{jk}+b_k^i,\sigma^i_{k}\right)}{\underset{k=1}{\overset{K_i}{\sum}}\pi^i_k\mathcal N\left(X_{ij}\ \big|\ \left(\mathbf w^i_{jk}\right)^T\mathbf p^i_{jk}+b_k^i,\sigma^i_{k}\right)}$.
\end{Lemma}

We can find that the equation of biases and variances replace the $\boldsymbol\mu$ in equation (2) with the linear function. The proof of these equations can be found in \textbf{Appendix}. However the weights $\mathbf w^i_k$ can not be obtained in the same way since they multiply by the value of maximal parental clique which is related to the data set. The restriction 
$\underset{k=1}{\overset{K_i}{\sum}} \pi_k^i=1$ is a hard constraint(\cite{boyd2004convex}) which is required to make equation (3) a distribution so the gradient descent optimization can not optimize equation (4). We may use softmax function on $\pi^i_k$ for k=1,...,$K_i$, or normalize each of $\pi^i_k$ by dividing sum of them after certain epochs, but this can not guarantee the local optimum. So we propose the double iteration optimization(DIO), \textbf{Algorithm 2} gives details of it.

\begin{algorithm}[htpb]
\renewcommand{\algorithmicrequire}{\textbf{Input:}}
\renewcommand{\algorithmicensure}{\textbf{Output:}}
\caption{Double Iteration Optimization}
\label{alg2}
\begin{algorithmic}[1]
\REQUIRE $\mathcal G$, $\mathbf D$, $inner\ iterations$, $outer\ iterations$
\STATE $outer\ epoch=0$, $inner\ epoch=0$;
\STATE $N$ is the number of instances in $\mathbf D$;
\WHILE{$outer\ epoch<outer\ iterations$}
\STATE $outer\ epoch=outer\ epoch+1$;
\FOR{i in nodes of $\mathcal G$}
\FOR{k in indexes of maximal parental clique of node i}
\STATE $N^i_k=0$;
\FOR{j in $\mathbf D$}
\STATE compute the $\gamma^i_{jk}$ by the equation in \textbf{Lemma 1};
\STATE $N^i_k=N^i_k+\gamma^i_{jk}$;
\ENDFOR
\STATE $\pi^i_k=\displaystyle\frac{N^i_k}{N}$;
\ENDFOR
\ENDFOR
\WHILE{$inner\ epoch<inner iterations$}
\STATE $inner\ epoch=inner\ epoch+1$;
\STATE use the mini-batch gradient descent optimization to update  $\mathbf w^i_k$, $b^i_k$, $\sigma^i_k$ to minimize equation (4) for k=1,...$K_i$, i=1,...n;
\ENDWHILE
\ENDWHILE
\ENSURE $\pi^i_k$, $\mathbf w^i_k$, $b^i_k$, $\sigma^i_k$ for k=1,...$K_i$, i=1,...n.
\end{algorithmic}
\end{algorithm}

\begin{figure}[h]
\centering
\includegraphics[scale=0.6]{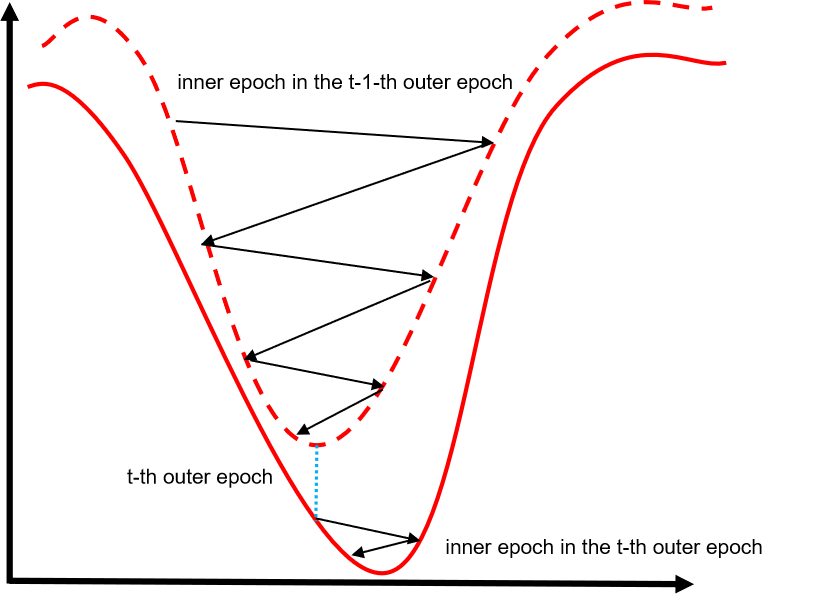}
\caption{The process of DIO:The red dashed curve is the loss function with respect to weights in t-1-th outer epoch. In inner iteration of t-1-th outer epoch, the gradient descent method(black arrows) finds the local optimum. After entering the next outer epoch, the blue dotted line represents the update of coefficient by EM method, the loss curve in t-th outer epoch(red solid curve) becomes deeper, then the gradient descent method reaches a deeper local optimum.}\label{fig1:vae}
\end{figure}

An outer epoch in DIO has two part, DIO uses the equation (7) to compute the coefficient $\pi^i_k$ in the first part, then fixes them and uses any kinds of gradient descent optimization to minimize the loss function in the second part in line 17, we use the mini-batch technique to accelerate the process.

\begin{Theorem} The double iteration optimization has hill-climbing property, if the inner epoch is big enough, then it will reach the local optimum.
\end{Theorem}

The latter part of \textbf{Theorem 3} is trivial since DIO reaches a local optimum
after every outer epoch for a big enough inner iteration. But in first part of outer epoch, it updates all the coefficients in last outer epoch, because of its hill-climbing property, the loss function reaches a smaller value than the local optimum at the end of last outer epoch, it can reaches a much smaller local optimum at the end of this outer epoch(figure 2). The complete proof is in \textbf{Appendix}

Notice that the mean and variance of Gaussian distribution can be more general form $\mathcal N(X\ |\ f(\boldsymbol\theta,\mathbf{Pa}_X),g(\boldsymbol\eta, \sigma))$ like in VAE, the weights in mixture model of this form is completely intractable, but we still can use DIO in this case. Moreover, we can not only update the coefficient in first part of outer epoch, but also update the biases and variances if they can be derived by the similar way in equation (6).

The drawback of DIO is obvious. In line 8, we need to use all the data set to update the coefficient which needs a lot of time and resource. If we also update the variances or biases in first part of outer epoch, it costs much more resource since variances and biases are more complicated to compute.
Although in each outer epoch, the loss can reach a smaller local optimum, it can not reach the global optimum after enough outer epochs, the update of coefficients only makes the current local optimum 'deeper' with respect to the weights, not finds a another smaller local optimum(Figure 2).

\section{Experiments}
In this section, we make some numerical experiments to exhibit the effectiveness of GMM-MPC. Three publicly-available data sets, Mental health (\cite{mhaelth}), House (\cite{timmurphy.org}) and Sachs (\cite{sachs2005causal}), are used. The basic information about them are given in Table \ref{tab:data}. Here, we are only concerned with continuous features, and the discrete features are removed from data sets. We do experiments on CPU i7-12700H, and all codes and part of data sets are provided in the Supplemental material. 

\begin{table*}[htpb]
\centering
\fontsize{8.5}{10}\selectfont
\caption{Basic information about three data sets}\label{tab:data}
\setlength{\tabcolsep}{3mm}{
\begin{tabular}{cccccccc}
\bottomrule[1.5pt]
\multirow{2}{*}{Data sets} & Continuous & Number of & Mini-batch \cr
  & Features & Instances & Size \cr
\bottomrule
\textbf{Mental health} & 12 & 125000 & 10000\cr
\textbf{House}         & 9  & 20640  & 7000 \cr
\textbf{Sachs}         & 11 & 7466   & 3000 \cr\bottomrule[1.5pt]
\end{tabular}}
\end{table*}

We first learn the graph structures of data sets. In this paper, three kinds of existing methods are directly borrowed, including PC (\cite{colombo2014order}), MMHC (\cite{tsamardinos2006max}) and greedy search (GS) (\cite{margaritis2003learning}) algorithms. The main purpose is to evaluate the performance of the proposed GMM-MPC under different graph structures. After performing normalization, we train the graph structure of every data set, and the results are exhibited in Figures 4, 5 and 6 in Supplement. We also present the corresponding number of edges in each graph structure in Tables \ref{tab:MH}, \ref{tab:H} and \ref{tab:S}. Based on these results, \textbf{Algorithms 1} and \textbf{2} are further applied to find MPCs for every node and optimize the model parameters, respectively. In the training process, Adam (\cite{kingma2014adam}) is selected as the gradient descent method for DIO, and the learning rate is set as $0.005$. The initial weights, bias and all of variances are set as 0, 0 and 1, respectively. We use $5$-fold cross validation to perform training on each 
data set. The training stops just before overfitting, which is implemented by an early stopping technique. This technique allows model testing to be done simultaneously after a certain amount of trainings, and identifies overfitting through observing the testing accuracy change. The overfitting corresponds to the time at which the testing accuracy begin to become low. At the moment, the number of training is called epochs. We use the average minus log-likehood as a criterion to evaluate the testing accuracy, and write down epochs in the form of $'inner\ iterations\times outer\ iterations'$ and testing accuracy in the form of $'mean\pm variance'$, with the results also reported in Tables \ref{tab:MH}, \ref{tab:H} and \ref{tab:S}. Additionally, considering the increasing parameters in introducing mixture models, we select the Bayesian information criterion (BIC)(\cite{neath2012bayesian}) to balance the model accuracy and complexity, defined by
\begin{align}
\text{BIC:}\ -L(\boldsymbol\theta,\mathbf D)+\frac12p\ln N.
\end{align}
Here, $L$ is the log-likelihood function, $\boldsymbol\theta$ represents parameters, $p$ is the number of model parameters and $N$ is the number of testing instances. The results of BIC are provided in those three tables as well.


\begin{table*}[htpb]
\centering
\fontsize{8.5}{10}\selectfont
\caption{Comparisons of different models on Mental health data set}\label{tab:MH}
\setlength{\tabcolsep}{3mm}{
\begin{tabular}{cccccccc}
\bottomrule[1.5pt]
\multirow{2}{*}{Model} & Structure & Number of & \multirow{2}{*}{Epochs} & Number of  & Average Minus & \multirow{2}{*}{BIC}  \cr
& Learning & Edges & &  Parameters & Log-Likelihood \cr
\bottomrule
 LG      & PC     & 11   & 200           & 35 & 18.25$\pm$0.056            & 456496$\pm$1287 \cr
 GMM     & PC     & 11   & 20$\times$6   & 99 & 17.96$\pm$0.04             & 449515$\pm$914\cr
 GMM-MPC & PC     & 11   & 20$\times$4   & 49 & \textbf{17.90$\pm$0.03}    & \textbf{447810$\pm$69} \cr 
 LG      & MMHC   & 32   & 320           & 56 & 17.18$\pm$0.13             & 429798$\pm$3062 \cr
 GMM     & MMHC   & 32   & 20$\times$7   & 190& 17.48$\pm$0.41             & 438028$\pm$10144 \cr
 GMM-MPC & MMHC   & 32   & 20$\times$6   & 115& \textbf{17.06$\pm$0.02}    & \textbf{427071$\pm$56}  \cr
 LG      & GS     & 31   & 240           & 55 & 17.71$\pm$0.04             & 442827$\pm$1039  \cr
 GMM     & GS     & 31   & 20$\times$7   & 194& 18.38$\pm$0.08             & 460551$\pm$1903 \cr
 GMM-MPC & GS     & 31   & 20$\times$4   & 108& \textbf{17.40$\pm$0.02}    & \textbf{435500$\pm$26} \cr\bottomrule[1.5pt]
\end{tabular}}
\end{table*}

\begin{table*}[htpb]
\centering
\fontsize{8.5}{10}\selectfont
\caption{Comparisons of different models on House data set}\label{tab:H}
\setlength{\tabcolsep}{2.5mm}{
\begin{tabular}{cccccccc}
\bottomrule[1.5pt]
\multirow{2}{*}{Model} & Structure & \multirow{2}{*}{Function}& Number of & \multirow{2}{*}{Epochs} & Number of & Likelihood & \multirow{2}{*}{BIC}  \cr
& Learning & & Edges & & Parameters & Score \cr
\bottomrule
 LG       & PC     & Linear  & 6   & 100          & 24 & 14.21$\pm$0.26            & 58753$\pm$1055\cr
 GMM      & PC     & Linear  & 6   & 15$\times$10 & 71 & 11.66$\pm$2.10            & 48430$\pm$8667\cr
 GMM-MPC  & PC     & Linear  & 6   & 15$\times$10 & 28 & \textbf{10.41$\pm$0.24}   & \textbf{43084$\pm$977} \cr 
 LG       & MMHC   & Linear  & 18  & 110          & 36 & 15.45$\pm$0.09            & 63947$\pm$387 \cr
 GMM      & MMHC   & Linear  & 18  & 15$\times$10 & 121& 10.94$\pm$1.74            & 45401$\pm$7186\cr
 GMM-MPC  & MMHC   & Linear  & 18  & 15$\times$10 & 53 & \textbf{10.25$\pm$1.19}   & \textbf{42551$\pm$4893} \cr
 LG       & GS     & Linear  & 15  & 120          & 33 & 16.67$\pm$0.69            & 68941$\pm$2858 \cr
 GMM      & GS     & Linear  & 15  & 15$\times$10 & 112& 11.14$\pm$3.03            & 46441$\pm$12503\cr
 GMM-MPC  & GS     & Linear  & 15  & 15$\times$10 & 58 & \textbf{10.55$\pm$2.32}   & \textbf{43783$\pm$9556}  \cr
 LG       & PC     & Sigmoid & 6   & 70           & 24 & 13.16$\pm$0.02            & 54432$\pm$98 \cr
 GMM-MPC & PC     & Sigmoid & 6   & 20$\times$10 & 28 & \textbf{10.43$\pm$1.50}        & \textbf{43190$\pm$6173}  \cr
 LG       & MMHC   & Sigmoid & 18  & 70           & 36 & 13.17$\pm$0.07            & 54522$\pm$276 \cr  
 GMM-MPC  & MMHC   & Sigmoid & 18  & 20$\times$10 & 53 & \textbf{10.30$\pm$1.48}   & \textbf{42745$\pm$6117} \cr
 LG       & GS     & Sigmoid & 15  & 70           & 33 & 13.18$\pm$0.06                 & 54530$\pm$249  \cr
 GMM-MPC  & GS     & Sigmoid & 15  & 20$\times$10 & 58 & \textbf{10.26$\pm$1.54}        & \textbf{42599$\pm$6373} \cr\bottomrule[1.5pt]
\end{tabular}}
\end{table*}

\begin{table*}[htpb]
\centering
\fontsize{8.5}{10}\selectfont
\caption{Comparisons of different models on Sachs data set}\label{tab:S}
\setlength{\tabcolsep}{2.5mm}{
\begin{tabular}{cccccccc}
\bottomrule[1.5pt]
\multirow{2}{*}{Model} & Structure & \multirow{2}{*}{Function}& Number of & \multirow{2}{*}{Epochs} & Number of & Likelihood & \multirow{2}{*}{BIC}  \cr
& Learning & & Edges & & Parameters & Score \cr
\bottomrule
 LG       & PC     & Linear  & 10  & 90           & 32  & 16.83$\pm$1.19            & 25250$\pm$1771 \cr
 GMM      & PC     & Linear  & 10  & 20$\times$5  & 94  & 14.09$\pm$0.29            & 21378$\pm$430  \cr
 GMM-MPC & PC     & Linear  & 10  & 20$\times$4  & 37  & \textbf{13.22$\pm$0.95}   & \textbf{19878$\pm$1421} \cr 
 LG       & MMHC   & Linear  & 27  & 100          & 49  & 22.58$\pm$1.00            & 33889$\pm$1498 \cr
 GMM      & MMHC   & Linear  & 27  & 20$\times$5  & 166 & 15.66$\pm$2.77            & 23987$\pm$4141 \cr 
 GMM-MPC  & MMHC   & Linear  & 27  & 20$\times$5  & 68  & \textbf{13.90$\pm$1.42}   & \textbf{21007$\pm$2125} \cr
 LG       & GS     & Linear  & 31  & 90           & 53  & 21.56$\pm$0.76            & 32379$\pm$1131 \cr
 GMM      & GS     & Linear  & 31  & 20$\times$5  & 178 & 13.87$\pm$1.19            & 21352$\pm$1773 \cr
 GMM-MPC  & GS     & Linear  & 31  & 20$\times$5  & 118 & \textbf{12.25$\pm$0.91}   & \textbf{18719$\pm$1365}  \cr
 LG       & PC     & Sigmoid & 10  & 60           & 32 & 15.79$\pm$0.27             & 23688$\pm$402  \cr
 GMM-MPC  & PC     & Sigmoid & 10  & 20$\times$4  & 37 & \textbf{13.21$\pm$0.50}    & \textbf{19853$\pm$747 }\cr
 LG       & MMHC   & Sigmoid & 27  & 60           & 49 & 15.64$\pm$0.45             & 23531$\pm$668 \cr
 GMM-MPC  & MMHC   & Sigmoid & 27  & 20$\times$5  & 68 & \textbf{12.84$\pm$0.94}    & \textbf{19425$\pm$1407} \cr
 LG       & GS     & Sigmoid & 31  & 50           & 53 & 15.59$\pm$0.28             & 23471$\pm$416  \cr
 GMM-MPC & GS     & Sigmoid & 31  & 20$\times$5  & 118 & \textbf{12.81$\pm$0.62}  & \textbf{19563$\pm$934}  \cr\bottomrule[1.5pt]
\end{tabular}}
\end{table*}

\begin{figure}[H]
\centering
\subfigure[]
{
    \begin{minipage}[b]{0.48\linewidth}
        \centering
        \includegraphics[scale=0.5]{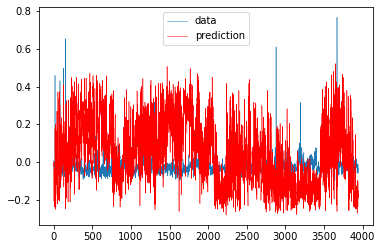}
    \end{minipage}
}
\subfigure[]
{
 	\begin{minipage}[b]{0.48\linewidth}
        \centering
        \includegraphics[scale=0.5]{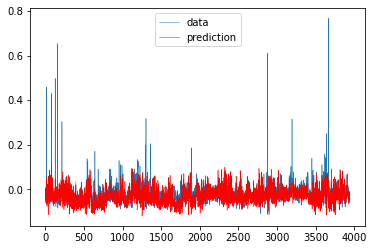}
    \end{minipage}
}
\caption{The comparison between test data in \textbf{House} data set and prediction. The feature 'Average Occupation' is the target node $T$. $T$ is a collider and its parents are 'Population' and 'Median House Value'. We use its parents to generate the value of $T$. The test data is normalized. (a)$T$ predicted by linear Gaussian with 'Sigmoid' term under GS.(b) $T$ predicted by GMM-MPC with 'Sigmoid' term under GS. }
\end{figure}

To fully exhibit the effectiveness of GMM-MPC, we simultaneously select another two models, linear Gaussian (LG) model and ordinary GMM \cite{roos2017dynamic,monti2013bayesian,liu2012bayesian} where each node is set to have the same number of branches, 3, in all experiments, for comparisons. The corresponding experimental results are also shown in Tables \ref{tab:MH}, \ref{tab:H} and \ref{tab:S}. 
For the latter two data sets, we compare three kinds of models. Moreover, we change the mean of Gaussian distributions with more complicated function (from Linear function to Sigmoid function $f(x)=\displaystyle\frac1{1+e^{-x}}$) to indicate that GMM-MPC can also have improvement on more general form of distribution, i.e., $P(X_i\ |\ \mathbf{Pa}_i)=\underset{k=1}{\overset{K_i}{\sum}}\pi^i_k\mathcal N\left(X_i\ \big|\ f\left(\left(\mathbf w^i_k\right)^T\mathbf p^i_{jk}\right)+b^i_k,\sigma^i_k\right).$ As can be seen from these tables, GMM-MPC completely outperform the other two models according to average testing accuracy and BIC whatever the graph structure (PC, MMHC or GS) is and whatever the function (linear or Sigmoid one) is. We also notice that ordinary GMM has much more parameters than GMM-MPC, but still has worse performance than GMM-MPC.
According to the \textbf{Algorithm 2}, the number of increased parameters is $O(kn)$ with $k$ to represent the number of maximal parental cliques of $T$. However in ordinary GMM, it is $\Theta(Kn)$ where $K$ is a hyperparameter. Figure 3 displays the prediction results of GMM-MPC and LG models with Sigmoid function on House data set. The former exhibit much more prediction power than the latter.


\section{Conclusion}
We use the mixture model on the Bayesian network based on the maximal parental cliques and propose a new optimization method DIO to find the local optimum of loss function. GMM-MPC can increase the generative and discriminative ability dramatically as we see in the tables and figure 3. More importantly, we can understand the meaning of every parameters i.e. the weights decide the influence of parents of nodes, the coefficients show how each cluster influences the prediction of target node, thus the iterpretability of BN remains. GMM-MPC may increase the number of parameters compare to LG, but this increase is acceptable under the concept of BIC and it has much less parameters than GMM. However in the training process, vanishing gradient and exploding gradient may appear, because unlike the log-likelihood function of LG, equation (4) puts mixture models in the $\ln$. Also in tables, we can find training GMM-MPC in BN always needs more epochs, e.g. in table 2, we train a LG with 'Sigmoid' only needs 70 epochs, but we need 200 epochs to train a GMM-MPC with 'Sigmoid'. These may be some points of our future efforts.


\bibliographystyle{unsrt}
\bibliography{reference}



\newpage\appendix

\section{Proofs of Theorems}

\subsection{Proof of Theorem 1}

To prove the mixture distributions (3) are all conjugate distributions for i=1,2,..n, we need to prove the prior distributions $P(X_i)$ and posterior distributions $P(X_i\ |\ \mathbf{Pa}_i)$ are in the same family. Since we define the posterior in (3), we need to obtain the prior distributions of $X_i$, i=1,2...n.

In DAG $\mathcal G$, there must be a node $X\in\mathcal V$ has no parent, otherwise $\mathcal G$ must have a cycle. So the prior and posterior distribution of $X$ are the form of $\overline P$, $\overline P$ can be seen as a mixture distribution whose number of component is one. By using induction method, for posterior distribution of any node $X_v\in \mathcal V$, we assume the joint distribution $P(\mathbf {Pa}_v)$ is a mixture distribution i.e.
\begin{align*} P(\mathbf{Pa})=\sum_{k=1}^{K_v}\pi^v_k\overline P^v_k(\mathbf{Pa}_v).
\end{align*} 

Then we obtain the joint distribution of $X_v$ and $\mathbf{Pa}_v$
\begin{align*} P_v(X_v, \mathbf{Pa}_v)&
=P_v(X_v\ |\ \mathbf{Pa}_v)P(\mathbf{Pa}_v)\\&
=\sum_{k=1}^{K_v}\pi^v_k\overline P^v_k(X_v\ |\ \mathbf{Pa}_v)\sum^{K_v}_{s=1}\eta^v_s\overline P^v_s(\mathbf{Pa}_v)\\&
=\sum_{k,s}\pi^v_k\eta^v_s\overline P^v_k(X_v\ |\ \mathbf{Pa}_v)\overline P^v_s(\mathbf{Pa}_v)\\&
=\sum_{k,s}\pi^v_k\eta^v_s\overline P^v_{ks}(X_v,\mathbf{Pa}_v)
\end{align*}

The last step is using the assumption of theorem that the component distribution $\overline P$ is conjugate distribution.Then by eliminating the variables $\mathbf{Pa}_v$, we have the prior distribution of $X_v$
\begin{align*} P(X_v)&
=\sum_{pa\in Val(\mathbf{Pa}_v)} P(X_v,pa)\\&
=\sum_{k,s}\pi^v_k\eta^v_s\sum_{pa\in Val(\mathbf{Pa}_v)} \overline P^v_{ks}(X_v,pa)\\&
=\sum_{k,s}\pi^v_k\eta^v_s\overline P^v_{ks}(X_v),
\end{align*}
where $Val(\mathbf{Pa}_v)$ is the set of all possible values of features $\mathbf{Pa}_v$. From equation above, the $\overline P^v_{ks}(X_v)$ is the form the $\overline P$ because of the conjugate of $\overline P$, and $\sum_{k,s}\pi^v_s\eta^v_s=1$, so the prior distribution $P(X_V)$ and the posterior distribution $P(X_v\ |\ \mathbf{Pa}_v)$ are in the same family of distribution, then the mixture distribution is conjugate.Moreover, 
\begin{align*}P(\mathcal V)&
=\prod_{i=1}^n P(X_i\ |\ \mathbf{Pa}_i)\\&
=\prod_{i=1}^n\sum_{k_i=1}^{K_i}\pi^i_{k_i}\overline P^i_{k_i}(X_i\ |\ \mathbf{Pa}_i)\\&
=\sum_{k_1,...k_n}\left(\prod^n_{i=1}\pi^i_{k_i}\right)\left(\prod^n_{i=1}\overline P^i_{k_i}(X_i\ |\ \mathbf{Pa}_i)\right)\\&
=\sum_{k_1,...k_n}\left(\prod^n_{i=1}\pi^i_{k_i}\right)\left(\overline P_{k_1,...k_n}(\mathcal V)\right)
\end{align*}
where $\overline P_{k_1,...k_n}(\mathcal G)=\overset{n}{\underset{i=1}{\prod}}\overline P^i_{k_i}(X_i\ |\ \mathbf{Pa}_i)$, and $\underset{k_1,...k_n}{\sum}\overset{n}{\underset{i=1}{\prod}}\pi^i_{k_i}=1$ so the joint distribution of graph $\mathcal G$ is mixture model. For any subset of nodes $\mathcal S\subset\mathcal V$, eliminating the rest of nodes $\mathcal V\backslash\mathcal S$ in $P(\mathcal V)$, we know the joint distribution of $\mathcal S$ $P(\mathcal S)$ is the form of $P(\mathcal V)$
$\hfill\blacksquare$

\subsection{Proof of Theorem 2}
Let us assume node $T\in \mathcal V$ is a collider, by the definition of collider, we can find two parents of $T$: $X$,$Y$, and $X\to Y\notin\mathcal E$ and $X\gets Y\notin\mathcal E$. Find any maximal clique $C_X$ that contains $X$, and any maximal clique $C_Y$ that contains Y, let $\mathbf{Pa}_1=\mathbf{Pa}_X\cap C_X$ and $\mathbf{Pa}_2=\mathbf{Pa}_Y\cap C_Y$, then $X\notin \mathbf{Pa}_2$, $Y\notin \mathbf{Pa}_1$,  $\mathbf{Pa}_1$ and $\mathbf {Pa}_2$ are two different maximal parental cliques. The coefficient of these two maximal parental cliques $\pi_1$, $\pi_2$ are non-zero, otherwise if $\pi_1=0$, then $X$ is not parent of $T$, since we choice $X$ and $Y$ arbitrarily, then $T$ is not a collider, this is contradictory to the assumption.

According to the \textbf{Theorem 1}, we know the joint distribution of $\mathbf{Pa}_i$ is mixture distribution  $P(\mathbf{Pa}_T)=\overset{K}{\underset{s=1}{\sum}}\eta_s\overline P^T_s(\mathbf{Pa}_T)$, let the conditional distribution of $T$ be $P(T\ |\ \mathbf{Pa}_T)=\overset{K}{\underset{k=1}{\sum}}\pi_k\overline P^T_k(T)$, then the marginal distribution of $T$ is 
\begin{align*} P(T)=\sum_{k,s}\pi_k\eta_s\overline P^T_{k,s}(T)
\end{align*}
There is at least one $\eta_s$ such that $\eta_s\neq0$, then in the equation above, $\pi_1\eta_s$ and $\pi_2\eta_2$ are non-zero, so the marginal distribution of $T$ has more than one component.

If $T$ is the descendant of collider, without loss of generality we can assume one of the node $Z\in\mathbf{Pa}_T$ is collider. From the proof of \textbf{Theorem 1}, where $m<n$, we know
\begin{align*} P(\mathbf{Pa}_T)&=\sum_{\mathcal V\backslash\mathcal S}P(\mathcal V)
=\sum_{\mathcal V\backslash\mathcal S}\sum_{k_1,...k_n}\left(\prod^n_{i=1}\pi^i_{k_i}\right)\left(\overline P_{k_1,...k_n}(\mathcal V)\right)\\&
=\sum_{k_1,...k_n}\left(\prod^n_{i=1}\pi^i_{k_i}\right)\left(\sum_{\mathcal V\backslash\mathcal S}\overline P_{k_1,...k_n}(\mathcal V)\right)\\&
=\sum_{k_1,...k_n}\left(\prod^n_{i=1}\pi^i_{k_i}\right)\left(\overline P_{k_1,...k_n}(\mathcal S)\right).
\end{align*}
We assume $X_1=Z$, then by the proof before, we know conditional distribution $P(Z\ |\ \mathbf{Pa}_Z)=\overset{K_1}{\underset{k_1=1}{\sum}}\pi^1_{k_1}\overline P^1_{k_1}(Z\ |\ \mathbf{Pa}^1_{k_1})$
has more than one component, without loss of generality, we assume $\pi^1_1\neq0$ and $\pi^1_2\neq0$, then there are at least two non-zero coefficient in $\pi^1_1\overset{n}{\underset{i=2}{\prod}}\pi^i_{k_i}$ and $\pi^1_2\overset{n}{\underset{i=2}{\prod}}\pi^i_{k_i}$ for $k_i=1,2,...K_i$, $i=2,...n$. Similarly, the marginal distribution of $T$ is 
\begin{align*} P(T)=\sum_{k_1,...k_n}\sum_{k}\pi_k\left(\prod^n_{i=1}\pi^i_{k_i}\right)\overline P^T_{k,k_1,...k_n}(T).
\end{align*}
It simply replaces the $\eta_s$ with $\left(\prod^n_{i=1}\pi^i_{k_i}\right)$, then the marginal distribution of $T$ has at least two components.$\hfill\blacksquare$

\subsection{Proof of lemma 1}
To minimize equation (4), we need to compute the value of Lagrange multiplier $\lambda_i$, let $L$ stands for the equation (4), setting the derivatives of equation (4) with respect to $\pi^i_k$, we have
\begin{align*} 0=-\sum^N_{j=1}\frac{\mathcal N\left(X_{ij}\ |\ \left(\mathbf{w}^i_k\right)^T\mathbf{p}^i_{jk}+b^i_k,\sigma^i_k\right)}{\overset{K_i}{\underset{k=1}{\sum}}\pi^i_k\mathcal N\left(X_{ij}\ |\ \left(\mathbf{w}^i_k\right)^T\mathbf{p}^i_{jk}+b^i_k,\sigma^i_k\right)}+\lambda_i
\end{align*}
Multiply $\pi^i_k$ on both side and sum all the k from 1 to $K_i$,
\begin{align*} 0&=-\sum^{K_i}_{k=1}\sum^N_{j=1}\frac{\pi^i_k\mathcal N\left(X_{ij}\ \big|\ \left(\mathbf{w}^i_k\right)^T\mathbf{p}^i_{jk}+b^i_k,\sigma^i_k\right)}{\overset{K_i}{\underset{k=1}{\sum}}\pi^i_k\mathcal N\left(X_{ij}\ \big|\ \left(\mathbf{w}^i_k\right)^T\mathbf{p}^i_{jk}+b^i_k,\sigma^i_k\right)}+\sum^{K_i}_{k=1}\lambda_i\pi^i_k \\&
=-\sum^{K_i}_{k=1}\sum^N_{j=1}\gamma^i_{jk}+\sum^{K_i}_{k=1}\lambda_i\pi^i_k\\&
=-N+\lambda_i,
\end{align*}
so $\lambda_i=N$ for $i=1,...n$.Thus 
\begin{align*} 0&=\frac{\partial L}{\partial\pi^i_k}=-\sum^N_{j=1}\frac{\pi^i_k\mathcal N\left(X_{ij}\ \big|\ \left(\mathbf{w}^i_k\right)^T\mathbf{p}^i_{jk}+b^i_k,\sigma^i_k\right)}{\overset{K_i}{\underset{k=1}{\sum}}\pi^i_k\mathcal N\left(X_{ij}\ \big|\ \left(\mathbf{w}^i_k\right)^T\mathbf{p}^i_{jk}+b^i_k,\sigma^i_k\right)}+N\pi^i_k \\&
=-N_k^i+N\pi^i_k
\end{align*}
which proves the equation (7). We can obtain the derivation of $L$ with respect to biases, variances and weights
\begin{align*}
0=\frac{\partial L}{\partial b^i_k}&=\sum^N_{j=1}\frac{\pi^i_k\mathcal N\left(X_{ij}\ \big|\ \left(\mathbf{w}^i_k\right)^T\mathbf{p}^i_{jk}+b^i_k,\sigma^i_k\right)}{\overset{K_i}{\underset{k=1}{\sum}}\pi^i_k\mathcal N\left(X_{ij}\ \big|\ \left(\mathbf{w}^i_k\right)^T\mathbf{p}^i_{jk}+b^i_k,\sigma^i_k\right)}\left(\frac{X_{ij}-\left(\mathbf w^i_k\right)^T\mathbf p^i_{jk}-b^i_k}{\sigma^i_k}\right)\\
\sum^N_{j=1}\gamma^i_{jk} b^i_k&=\sum^N_{j=1}\gamma^i_{jk}\left(X_{ij}-\left(\mathbf w^i_k\right)^T\mathbf p^i_{jk}\right)\\
b^i_k&=\frac{1}{N^i_k}\sum^N_{j=1}\gamma^i_{jk}\left(X_{ij}-(\mathbf w^i_k)^T\mathbf p^i_{jk}\right)
\end{align*}
\begin{align*}0=\frac{\partial L}{\partial\sigma^i_k}&=\sum^N_{j=1}\frac{\pi^i_k\mathcal N\left(X_{ij}\ \big|\ \left(\mathbf{w}^i_k\right)^T\mathbf{p}^i_{jk}+b^i_k,\sigma^i_k\right)}{\overset{K_i}{\underset{k=1}{\sum}}\pi^i_k\mathcal N\left(X_{ij}\ \big|\ \left(\mathbf{w}^i_k\right)^T\mathbf{p}^i_{jk}+b^i_k,\sigma^i_k\right)}\left(-\frac{\left(X_{ij}-\left(\mathbf w^i_k\right)^T\mathbf p^i_{jk}-b^i_k\right)^2}{2\left(\sigma^i_k\right)}+\frac{1}{2\sigma^i_n}\right)\\
\sum^N_{j=1}\gamma^i_{jk}\frac{1}{\sigma^i_n}&=\sum^N_{j=1}\gamma^i_{jk}\frac{\left(X_{ij}-\left(\mathbf w^i_k\right)^T\mathbf p^i_{jk}-b^i_k\right)^2}{\left(\sigma^i_k\right)}\\
\sigma^i_n&=\sum^N_{j=1}\gamma^i_{jk}\left(X_{ij}-\left(\mathbf w^i_k\right)^T\mathbf p^i_{jk}-b^i_k\right)^2.
\end{align*}
\begin{align*}0=\frac{\partial L}{\partial \mathbf w^i_k}&=-\sum^N_{j=1}\frac{\pi^i_k\mathcal N\left(X_{ij}\ \big|\ \left(\mathbf{w}^i_k\right)^T\mathbf{p}^i_{jk}+b^i_k,\sigma^i_k\right)}{\overset{K_i}{\underset{k=1}{\sum}}\pi^i_k\mathcal N\left(X_{ij}\ \big|\ \left(\mathbf{w}^i_k\right)^T\mathbf{p}^i_{jk}+b^i_k,\sigma^i_k\right)}\left(\frac{X_{ij}-\left(\mathbf w^i_k\right)^T\mathbf p^i_{jk}-b^i_k}{\sigma^i_k}\right)\left(\mathbf{p}^i_{jk}\right)\\
\sum^N_{j=1}\gamma^i_{jk}\left(\mathbf w^i_k\right)^T\mathbf p^i_{jk}\mathbf p^i_{jk}&=\sum^N_{j=1}\gamma^i_{jk}\left(X_{ij}-b^i_k\right)\left(\mathbf p^i_{jk}\right)\\
\sum^N_{j=1}\gamma^i_{jk}\left(\left(\mathbf p^i_{jk}\right)^T\mathbf p^i_{jk}\right)\left(\mathbf w^i_k\right)&=\sum^N_{j=1}\gamma^i_{jk}\left(X_{ij}-b^i_k\right)\left(\mathbf p^i_{jk}\right)\\
\end{align*}

These derive equation (5) (6)$\hfill\blacksquare$

\subsection{Proof of Theorem 3}
To prove the hill-climbing property of DIO, we need to prove after every outer epoch, the value 0f loss function (4) is become smaller. In one outer epoch, the inner iteration is gradient descent which is hill-climbing algorithm, we only need to prove the update of coefficients is hill-climbing.

Since in the first part of outer epoch, the weights, biases and variances are constant, we write $\mathcal N^i_{jk}=\mathcal N\left(X_{ij}\ \big|\ \left(\mathbf w^i_k\right)^T\mathbf p^i_{jk}+b^i_k,\sigma^i_k\right)$ for short. Let $\pi^i_k$ be the coefficients at present and $\tilde\pi^i_k$ be the coefficients in next outer epoch, then by the \textbf{Lemma 1},
\begin{align*} \tilde\pi^i_k=\frac{1}{N}\sum^N_{j=1}\frac{\pi^i_k\mathcal N^i_{jk}}{\overset{K_i}{\underset{k=1}{\sum}}\pi^i_k\mathcal N^i_{jk}}.
\end{align*}
So $\overset{K_i}{\underset{k=1}{\sum}}\tilde\pi^i_k=1$, the coefficients satisfy the constraint in every inner and outer epoch.

We use $L$ and $L'$ be the loss function at present and at next outer epoch, 
\begin{align*} L-L'&=\sum^n_{i=1}\sum^N_{j=1}\ln\left(\sum^{K_i}_{k=1}\tilde\pi^i_{jk}\mathcal N^i_{jk}\right)-\sum^n_{i=1}\sum^N_{j=1}\ln\left(\sum^{K_i}_{k=1}\pi^i_{jk}\mathcal N^i_{jk}\right)\\&
=\sum^n_{i=1}\sum^N_{j=1}\ln\left(\frac1N\sum^N_{j=1}\frac{\overset{K_i}{\underset{k=1}{\sum}}\pi^i_k\left(\mathcal N^i_{jk}\right)^2}{\overset{K_i}{\underset{k=1}{\sum}}\pi^i_k\mathcal N^i_{jk}}\right)-\sum^n_{i=1}\sum^N_{j=1}\ln\left(\sum^{K_i}_{k=1}\pi^i_{jk}\mathcal N^i_{jk}\right)\\&
=\sum^n_{i=1}\sum^N_{j=1}\ln\left(\frac1N\sum^N_{j=1}\frac{\overset{K_i}{\underset{k=1}{\sum}}\pi^i_k\left(\mathcal N^i_{jk}\right)^2}{\left(\overset{K_i}{\underset{k=1}{\sum}}\pi^i_k\mathcal N^i_{jk}\right)^2}\right),
\end{align*}
since $f(x)=x^2$ is convex then 
\begin{align*} \left(\sum^{K_i}_{k=1}\pi^i_k\mathcal N^i_{jk}\right)^2\leq \sum^{K_i}_{k=1}\pi^i_k\left(\mathcal N^i_{jk}\right)^2
\end{align*}
continue the process, we obtain 
\begin{align*}
L-L'&\geq\sum^n_{i=1}\sum^N_{j=1}\ln\left(\frac1N\sum^N_{j=1}1\right)=0
\end{align*}
$\hfill\blacksquare$

\section{Other details and results in experiments}
The details of experiments are shown below.

As we mentioned in the conclusion, GMM in BN may face the vanishing gradient and exploding gradient. To avoid these phenomenons, we add a constant $\epsilon$ in the $\ln$ in equation (4),
\begin{align*} -\sum^N_{j=1}\sum^n_{i=1}\ln\left(\sum^{K_i}_{k=1}\pi_k^i\mathcal N\left(X_{ij}\ \big|\ \left(\mathbf w_k^i\right)^T\mathbf p^i_{jk}+b_k^i,\sigma^i_{k}\right)+\epsilon\right).
\end{align*}
And add the $\epsilon$ in the update of coefficient in \textbf{Algorithm 2}, 
\begin{align*} \gamma^i_{jk}=\frac{\pi^i_k\mathcal N\left(X_{ij}\ \big|\ \left(\mathbf w_k^i\right)^T\mathbf p^i_{jk}+b_k^i,\sigma^i_{k}\right)}{\underset{k=1}{\overset{K_i}{\sum}}\pi^i_k\mathcal N\left(X_{ij}\ \big|\ \left(\mathbf w^i_{jk}\right)^T\mathbf p^i_{jk}+b_k^i,\sigma^i_{k}\right)+\epsilon}\ \ \ \ \ \  \pi^i_k=\frac1N\sum^{N}_{j=1}\gamma^i_{jk}
\end{align*}
In experiments, we set $\epsilon=1\times10^{-8}$. Although we apply this measure in experiment, we may still meet vanishing gradient when the training loss become too small, early stopping can completely solve this problem.
\begin{figure}[htpb]
\centering
\subfigure[]
{
    \begin{minipage}[b]{1\linewidth}
        \centering
        \includegraphics[scale=0.8]{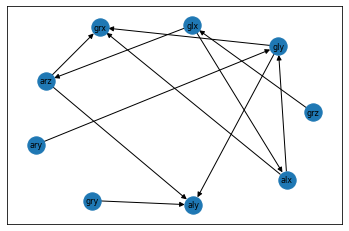}
    \end{minipage}
}
\subfigure[]
{
 	\begin{minipage}[b]{1\linewidth}
        \centering
        \includegraphics[scale=0.8]{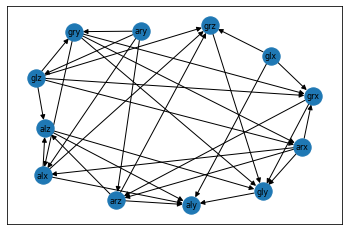}
    \end{minipage}
}
\subfigure[]
{
 	\begin{minipage}[b]{1\linewidth}
        \centering
        \includegraphics[scale=0.8]{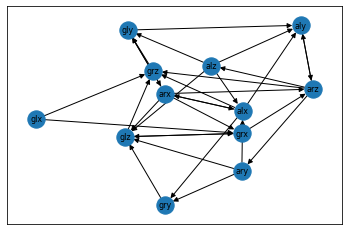}
    \end{minipage}
}

\caption{Bayesian network generated by different structure learning algorithms on \textbf{mental health} data set (a) PC ;(b) MMHC; (c) GS;}\label{MH_data}
\end{figure}

\begin{figure}[htpb]
\centering
\subfigure[]
{
    \begin{minipage}[b]{1\linewidth}
        \centering
        \includegraphics[scale=0.8]{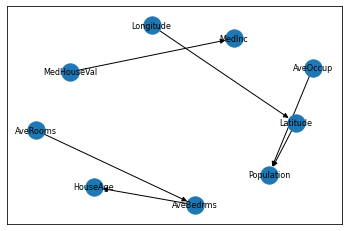}
    \end{minipage}
}
\subfigure[]
{
 	\begin{minipage}[b]{1\linewidth}
        \centering
        \includegraphics[scale=0.8]{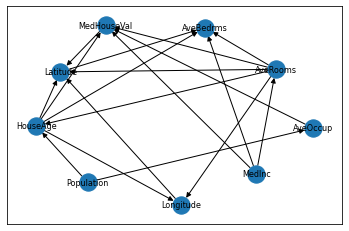}
    \end{minipage}
}
\subfigure[]
{
 	\begin{minipage}[b]{1\linewidth}
        \centering
        \includegraphics[scale=0.8]{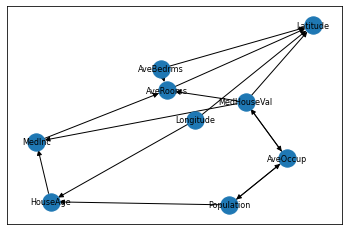}
    \end{minipage}
}

\caption{Bayesian network generated by different structure learning algorithms on \textbf{House} data set (a) PC ;(b) MMHC; (c) GS;}\label{H_data}
\end{figure}

\begin{figure}[htpb]
\centering
\subfigure[]
{
    \begin{minipage}[b]{1\linewidth}
        \centering
        \includegraphics[scale=0.8]{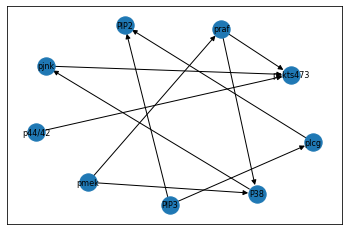}
    \end{minipage}
}
\subfigure[]
{
 	\begin{minipage}[b]{1\linewidth}
        \centering
        \includegraphics[scale=0.8]{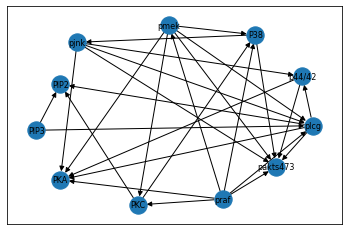}
    \end{minipage}
}
\subfigure[]
{
 	\begin{minipage}[b]{1\linewidth}
        \centering
        \includegraphics[scale=0.8]{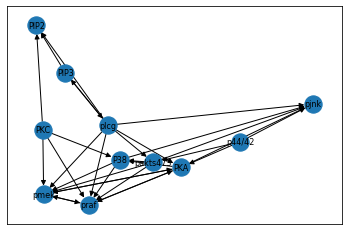}
    \end{minipage}
}

\caption{Bayesian network generated by different structure learning algorithms on \textbf{sachs} data set (a) PC ;(b) MMHC; (c) GS;}\label{S_data}
\end{figure}
\end{document}